\pdfoutput=1

\documentclass[11pt]{article}

\usepackage{EMNLP2023}

\usepackage{times}
\usepackage{latexsym}

\usepackage[T1]{fontenc}

\usepackage[utf8]{inputenc}

\usepackage{microtype}

\usepackage{inconsolata}

\usepackage{hyperref}       
\usepackage{url}            
\usepackage{booktabs}      
\usepackage{amsfonts}      
\usepackage{nicefrac}       
\usepackage{microtype}      
\usepackage{xcolor}        
\usepackage{amsmath}
\usepackage{amssymb}
\usepackage[section]{placeins}
\usepackage{graphicx}
\usepackage{natbib}
\usepackage{booktabs}
\usepackage{tabularx}
\usepackage{xcolor}
\usepackage{xspace}
\usepackage{colortbl} 
\usepackage{soul} 
\usepackage{tabularx}
\usepackage{stfloats}
\usepackage{algorithm,algorithmicx}
\usepackage[noend]{algpseudocode}
\usepackage{algcompatible}
\usepackage{pifont}
\usepackage{makecell}
\graphicspath{{./figures/}}
\usepackage{multirow}

\definecolor{lightgrey}{HTML}{dcdbdb}
\sethlcolor{lightgrey}
\definecolor{lightblue}{HTML}{E8F0FE}
\definecolor{lightblue}{HTML}{E8F0FE}
\definecolor{gray}{HTML}{9aa0a6}
\definecolor{lightpink}{HTML}{F48FB1}
\definecolor{lightred}{HTML}{FFCBC9}
\definecolor{lightcyan}{HTML}{80DEEA}
\newcommand{\cc}[0]{\cellcolor{lightblue}}
\newcommand{\tcc}[0]{\colorbox{lightblue}}
\newcommand{\bb}[0]{\cellcolor{lightgrey}}

\usepackage[skins]{tcolorbox} 
\tcbuselibrary{breakable} 
\newtcolorbox[auto counter, number within=section, list type=subsubsection, list inside=toc]{sectionbox}[2][]{
colback=white!98!gray, colframe=black, 
colbacktitle=white!90!gray, coltitle=black, 
fonttitle=\bfseries,
title={#2}, 
list entry={Comment \thetcbcounter\quad}
}

\usepackage{soul}

\newcommand{\blue}[1]{\textcolor{blue}{#1}}

\newcommand{\ourmodel}{G-LLaVA\xspace}
\newcommand{\ourdata}{Geo170K\xspace}

\title{%
    \ourmodel \raisebox{-0.25cm}{\includegraphics[width=1.5cm]{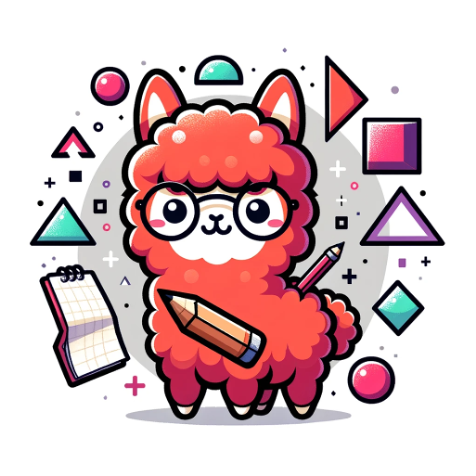}}: Solving Geometric Problem with Multi-Modal Large Language Model
}

\author{Jiahui Gao$^{1,2}$\thanks{\, Equal Contribution. 
}, Renjie Pi$^3$\footnotemark[1], 
\textbf{Jipeng Zhang}$^3$,
\textbf{Jiacheng Ye$^2$}, 
\textbf{Wanjun Zhong}$^1$,\\
\textbf{Yufei Wang$^1$,}
\textbf{Lanqing Hong$^1$,} 
\textbf{Jianhua Han$^1$,}
\textbf{Hang Xu$^1$,}\\
\textbf{Zhenguo Li$^1$,}
\textbf{Lingpeng Kong$^2$}
\\
$^1$Noah’s Ark Lab \quad 
$^2$The University of Hong Kong \\
$^3$The Hong Kong University of Science and Technology
\\
\texttt{\small sumiler@connect.hku.hk, rpi@connect.ust.hk}
}
\begin{document}

\maketitle
\vspace{5mm}  

Large language models (LLMs) have shown remarkable proficiency in human-level reasoning and generation capabilities, which encourages extensive research on their application in mathematical problem solving. However, current work has been largely focused on text-based mathematical problems, with limited investigation in problems involving geometric information. Addressing this gap, we aim to enable LLMs to solve geometric problems by understanding image input. We first analyze the limitations of current Multimodal Large Language Models (MLLMs) in this area: they struggle to accurately comprehending basic geometric elements and their relationships.
To overcome these challenges, we take advantage of the unique characteristics of geometric problems (such as unique geometric logical form, and geometric scalability) and the capacity of the textual LLMs to build an enriched multimodal geometry dataset based on existing data. The augmented dataset, \ourdata, contains more than 170K geometric image-caption and question-answer pairs. 
Utilizing our constructed \ourdata dataset, we develop \ourmodel, which demonstrates exceptional performance in solving geometric problems, significantly outperforming GPT-4-V on the MathVista benchmark with only 7B parameters.
\section{Introduction}
Large language models (LLMs) exhibit human-like proficiency in reasoning~\citep{wei2022chain, wang2022self, zhou2022least} and generation~\citep{ouyang2022training,touvron2023llama}, which encourages extensive research on their application in mathematical problem solving~\citep{fu2023specializing,gou2023tora,yue2023mammoth,luo2023wizardmath,
zhao2023sego,zhao2023decomposing,jiang2023draft}. These problems often require highly sophisticated and symbolic reasoning capabilities, often considered impossible to solve before the era of LLMs.

It is an intuitive approach to use LLMs for mathematical reasoning problems presented in a textual form. Nevertheless, a substantial proportion of mathematical reasoning problems necessitate the comprehension of geometric information. Moreover, even when certain problems do not overtly pertain to geometric information on the surface, the integration of geometrical-based methods often holds significant practical implications (e.g., analytic number theory). With the advent of GPT-4V~\citep{openai2023gpt4}, Gemini\footnote{Gemini, a concurrent work, was released one week before our submission. Consequently, our work is primarily benchmarked against GPT4-V and other MLLMs.}~\citep{gemini2023}, and numerous multi-modal large language models (MLLMs) ~\citep{zhu2023minigpt4,liu2023llava,dai2023instructblip,li2023blip2,bai2023qwenvl,lai2023lisa,gao2023llamaadapter, pi2023perceptiongpt}, recent work has progressively looking into employing MLLMs to tackle geometric reasoning problems in mathematics~\citep{yang2023dawn,lu2023mathvista,yue2023mmmu}. 

\begin{figure*}[h]
\centering
\includegraphics[width=1\textwidth]{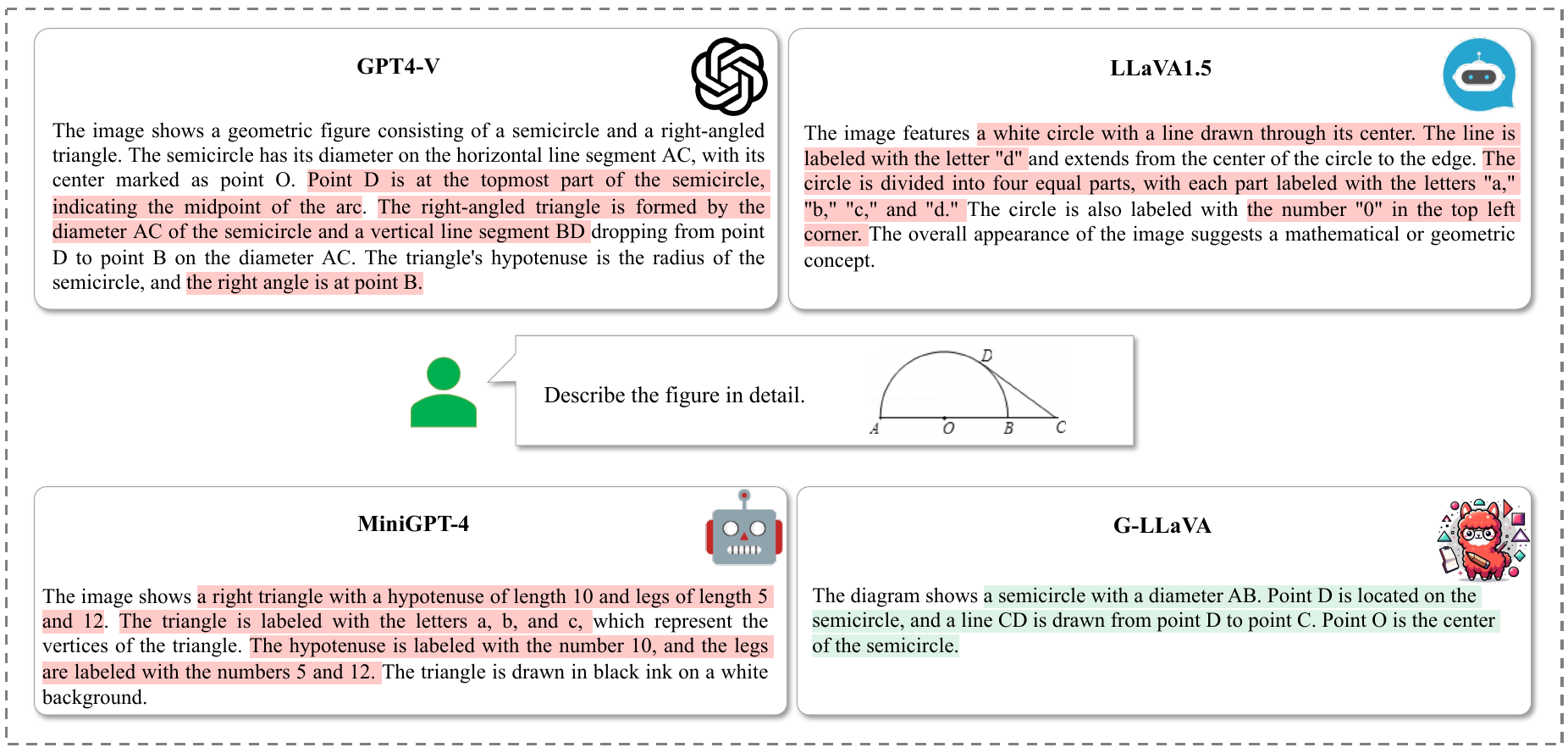} 
\vspace{-2mm}
\caption{ State-of-the-art MLLMs suffer severe \colorbox{lightred}{hallucination} on geometric figures, which greatly hinders their abilities on solving geometric problems. On the other hand, our \ourmodel's ability to interpret geometric figure is boosted after the alignment phase with our curated dataset. 
}\label{fig:comparison}
\end{figure*}

However, we have observed that even with the most advanced MLLMs, current systems still exhibit limitations in addressing geometric problems due to challenges in accurately comprehending geometric figures. 
For instance, as demonstrated in Figure~\ref{fig:comparison}, GPT4-V often produces inaccurate descriptions for geometric figures.  Specifically, the model struggles with understanding the relationships between fundamental elements like points and lines, and in accurately interpreting elements such as the degree of an angle.
We presume that the underlying reason for this may be the fact that these MLLMs are typically trained with images and descriptions from the general domain, and the ability to understand such semantics differs significantly from that required for geometric reasoning. 

To address this issue, one of the most direct and effective approaches is to enhance current MLLMs by augmenting them with data containing high-quality descriptions of geometric information~\citep{ye2022zerogen,meng2022generating}.
However, a significant challenge arises from the limited size of the largest publicly available geometric problem dataset, which contains only a few thousand question-answer pairs. 
Additionally, the current datasets lack descriptions of geometric images and exhibit a limited range of problem-solving methods, which constrains the model's ability to understand basic geometric elements and affect its problem-solving capabilities.

\begin{figure*}[t!]
\centering
\includegraphics[width=1\textwidth]{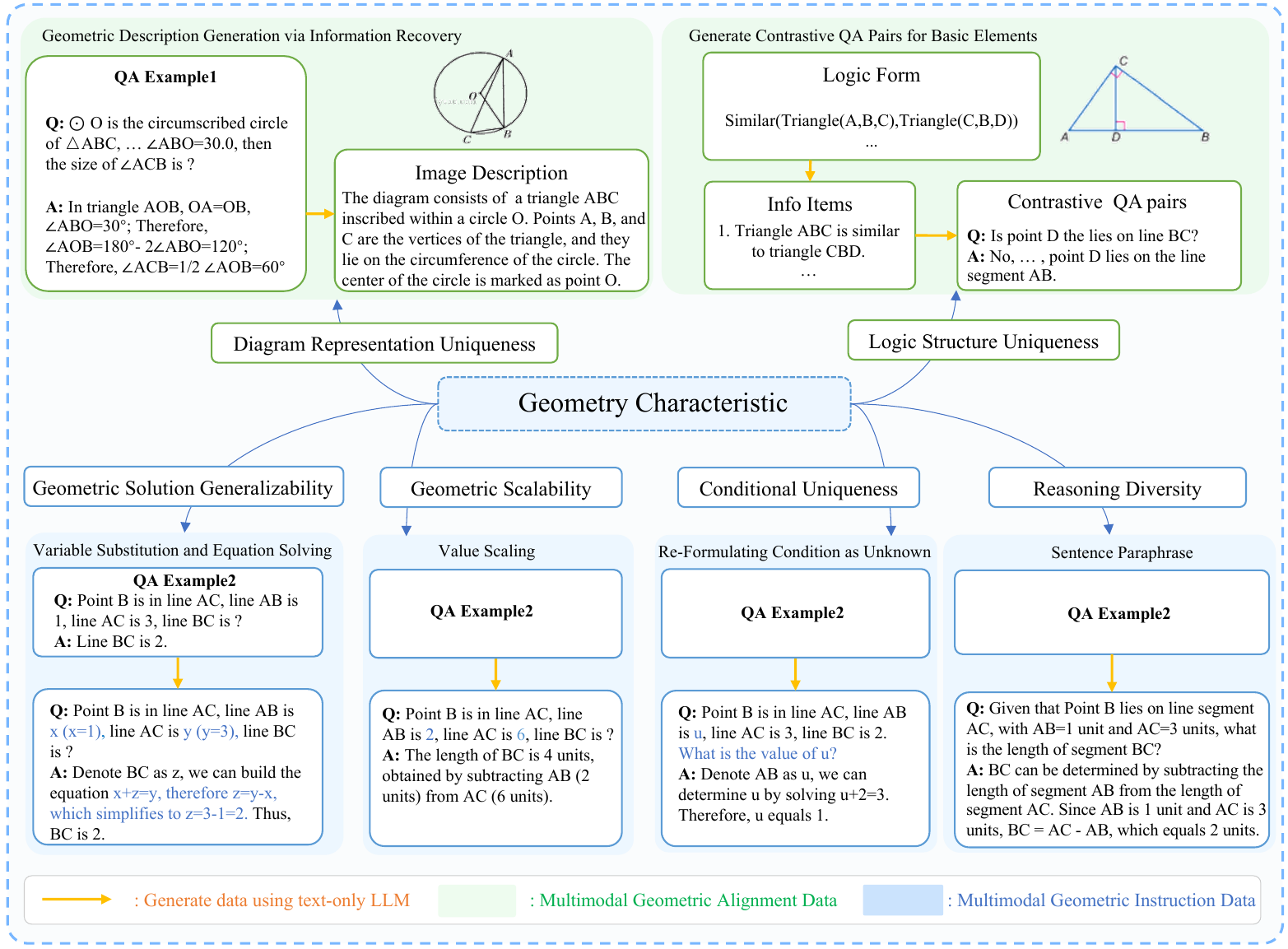} 
\vspace{-2mm}
\caption{Framework of our multi-modal geometric data generation using the characteristics of geometry problems. }\label{fig:main_figure}
\end{figure*}

In this paper, we propose to synthesize geometric visual-text data leveraging existing datasets via text-only LLMs (e.g., ChatGPT). 
More specifically, we utilize the geometry characteristic to construct a multi-modal geometry dataset, building upon existing datasets. The data generation process involves incorporating utilizing uniqueness of geometric logic form, geomertic representation uniqueness, geometric scalability, etc (as shown in Figure~\ref{fig:main_figure}).
We term our generated dataset \ourdata,  which contains around 60,000 geometric image-caption pairs and  more than 110,000 question-answer pairs. This dataset is 28 times larger than GeoQA+, greatly expanding the coverage of geometric problems. With our collected \ourdata, we derive \ourmodel, a MLLM capable of solving geometric problems,
surpassing SOTA MLLMs by a large margin.
Specifically, \ourmodel-13B outperforms LLaVA-13B by 27.4 on GPS minitest split of MathVista~\citep{lu2023mathvista}. In addition, with only \ourmodel-7B, it is able to surpass the powerful GPT4-V on the geometry problem solving questions. Code and data will be available at \url{https://github.com/pipilurj/G-LLaVA}.

\section{Related Work}
\paragraph{Multi-Modal Large Language Model.}
Recent years have witnessed transformative advancements in the development of large language models (LLMs), characterized by a series of pioneering studies~\citep{brown2020language, scao2022bloom, chowdhery2022palm, smith2022using, hoffmann2022training, ouyang2022training, touvron2023llama, bai2022training}. These breakthroughs have significantly elevated the capabilities of language understanding and generation, showcasing near-human proficiency across diverse tasks. Concurrently, the success of LLMs has inspired explorations into vision-language interaction, leading to the emergence of multi-modal large language models (MLLMs)~\citep{liu2023llava, li2023blip2, dai2023instructblip, zhu2023minigpt4, dai2023instructblip, openai2023gpt4, bai2023qwenvl, su2023pandagpt, gao2023llamaadapter}. These models have exhibited remarkable capabilities in synthesizing detailed descriptions and engaging in dialogue based on visual inputs. However, we observe that even the state-of-the-art MLLMs face challenges in resolving geometric problems using diagrams and figures.
\paragraph{Geometry Problem Solving.}
The Geometry problem reasoning is an challenging visual mathematical reasoning problem. Early efforts by \citet{seo2015solving, sachan2017textbooks, alvin2017synthesis, sachan2017learning} focused on creating datasets through manual efforts. More recent approaches have introduced enhanced methods and datasets, including Geometry3K \citep{lu2021inter}, GeoQA \citep{chen-etal-2021-geoqa}, GeoQA+ \citep{cao2022augmented}, UniGeo \citep{chen2022unigeo}, UniMath \citep{liang_unimath}, and SCA-GPS \citep{sca_gps}, aiming to improve both performance and explainability. However, the scale of current datasets remains limited, and the performance of traditional models in this domain has not achieved the level observed in other areas of mathematical problem solving, particularly when compared to methods that utilize large language models for solving math word problems \citep{gsm8k, wei2022chain, gou2023tora}.

\paragraph{Data Generation via LLM.} Bootstrapping data from pretrained models has long been an active area of research. \cite{ye2022zerogen, meng2022generating} generates training data using pretrained language models such as GPT-2 for classification tasks. \cite{gao2023selfguided} improves the quality of generated dataset via bi-level approach. \cite{ye-etal-2022-progen} utilizes influence function to select in-context examples to aid data generation. Recently, automatic data generation becomes more ubiquitous with the advent of powerful LLMs such as ChatGPT, a line of recent works utilize ChatGPT-generated data to perform instruction tuning~\citep{wang2023selfinstruct, peng2023instruction, alpaca, liu2023llava, zhu2023minigpt4, bai2023qwenvl, pi2023detgpt, su2023pandagpt, yu2023metamath, chen2023shikra, zhang2023gpt4roi}.
\section{Observation}
We observe that most state-of-the-art (SOTA) MLLMs, although being adept at understanding daily visual scenes, have difficulty in comprehending geometric figures, even if they are simple and straightforward for humans. In Figure~\ref{fig:comparison}, we demonstrate the descriptions generated by SOTA MLLMs for geometric figure. We observe that severe hallucination exists in all the generated descriptions. 

More specifically, we find GPT4-V has difficulty understanding relationships between basic elements like points and lines, and also struggles with precisely interpreting these elements themselves (such as the angle B in Figure~\ref{fig:comparison}). Furthermore, smaller MLLMs like LLaVA1.5 and MiniGPT4 demonstrate even greater difficulty in accurately identifying the types of geometric shapes present in a figure.

This inadequacy in interpreting geometric diagrams may be one of the major causes for the failure in solving geometric problems. In contrast, actual geometric diagrams typically exhibit clear and well-defined relationships among their elements. This geometry characteristic can be utilized to develop datasets that help mitigate the above issues and mitigate hallucination.

\begin{table*}
\centering
\begin{minipage}{1.0\textwidth}\vspace{0mm}    \centering
\begin{sectionbox}[]{Geometric Description Generation via Information Recovery }
    \centering
      \footnotesize
    \begin{tabular}{p{0.97\textwidth} c}
{ {\bf QA Pair:} } & \\
Question: As shown in the figure, circle O is the circumscribed circle of triangle ABC, and it is
& \hspace{-2.2cm} \multirow{5}{*}{ \includegraphics[height=2.0cm]{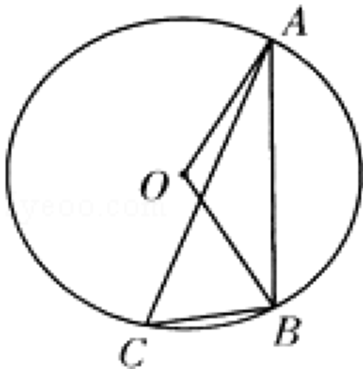}  }\\
known that angle ABO = 30.0, then the size of angle ACB is ()\\
Answer: In triangle AOB, OA=OB, angle ABO=30°; Therefore, angle AOB=180°- 2 angle ABO\\=120°; Therefore, angle ACB=1/2angle AOB=60°
\\
\rule{0.85\linewidth}{0.5pt} & \\
\bf Diagram Description:  & \\
\cc{The diagram consists of a triangle ABC inscribed within a circle, where the circle is denoted as circle O. Points A, B, and C are the vertices of the triangle, and they all lie on the circumference of the circle.  The center of the circle is marked as point O.}
\end{tabular}
\end{sectionbox}
\vspace{-5mm}
\caption{Full geometric diagram description generation via inverse information recovery. The description is generated based on the textual QA pair. The upper section shows the QA pair employed to instruct text-only ChatGPT, while the lower section (\tcc{in blue}) shows the responses produced by ChatGPT.}
    \label{tab:generate_caption_pretraining}
\end{minipage}
\end{table*}

\section{Geometric Data Generation}
While previous efforts have been made to address multi-modal geometry problems~\citep{chen-etal-2021-geoqa,chen2022unigeo,cao2022augmented}, the availability of geometry datasets remains limited.
The key limitations of existing datasets are threefold: (1) limited data volume (a few thousands for the largest dataset), (2) absence of detailed descriptions for geometric images, and (3) a lack of diversity in problem-solving methodologies and answer pathways. 
This limitation presents challenges for MLLMs in accurately understanding geometric elements and providing precise geometric solutions. 

To address this issue, we utilize the geometry characteristic to construct a multi-modal geometry dataset based upon existing dataset. This dataset includes two parts: an alignment dataset to provide MLLMs with fundamental geometric knowledge and an instruction-tuning dataset to improve the assistant's ability to understand user instructions and generate accurate geometry solutions.

\begin{table*}[t!]\centering
\begin{minipage}{1.0\textwidth}\vspace{0mm}    \centering
\begin{sectionbox}[]{Contrastive QA Pairs for Basic Elements}
    \centering
      \footnotesize
    \begin{tabular}{p{0.97\textwidth} c}
{ {\bf Logic Form:} } & \\
\texttt{Similar(Triangle(A,B,C),Triangle(C,B,D))}
 & \hspace{-3.2cm} \multirow{5}{*}{ \includegraphics[height=2.0cm]{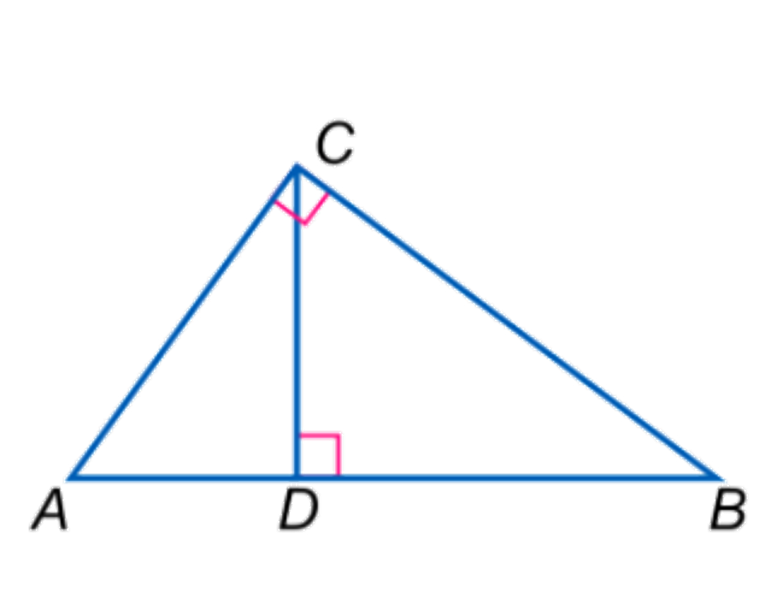}  }\\
\texttt{Triangle(A,B,C)}\\
\texttt{Triangle(A,C,B)}\\
\texttt{Equals(LengthOf(Line(C, B)), 5)}\\
\texttt{Equals(LengthOf(Line(C, D)), 3)}\\
\texttt{Equals(LengthOf(Line(B, D)), 4)}\\
\texttt{PointLiesOnLine(D, Line(A, B))}\\
\texttt{Perpendicular(Line(A, C), Line(C, B))}\\
\texttt{Perpendicular(Line(C, D), Line(A, D))}\\
\rule{0.8\linewidth}{0.5pt} & \\
\makecell[c]{\textit{Phase-1 Generation}}\\
{\bf Info Items:}  & \\
\cc{The geometric figure have the following information:}\\
\cc{Info 1: The figure represents a triangle with vertices A, B, and C.}\\
\cc{Info 2: Triangle ABC is similar to triangle CBD.}\\
\cc{Info 3: Point D lies on the line segment AB.}\\
\cc{Info 4: Line segment AC is perpendicular to line segment CB.}\\
\cc{...}\\
{\bf Summary: }  & \\
\cc{The figure depicts a triangle ABC and triangle CBD, which is similar to triangle ABC. Additionally, triangle ACB is also present in the figure. The lengths of the segments CB, CD, and BD are 5, 3, and 4 units respectively. Point D lies on the line segment AB, and perpendicularity is indicated between the line segments AC and CB, as well as between the line segments CD and AD.}
\\
\rule{0.8\linewidth}{0.5pt} & \\
\makecell[c]{\textit{Phase-2 Generation}}\\
{\bf QA pairs: }  & \\
\cc{1. Question: Does the diagram include a triangle formed by vertices A, B, and D?}\\
\cc{Answer: \textbf{No}, the diagram does not include a triangle formed by vertices A, B, and D. The triangles present are ABC and CBD, and the point D lies on the line segment AB.}\\
\cc{2. Question: Does the diagram contain a triangle with vertices A, C, and B?}\\
\cc{Answer: \textbf{Yes}, the diagram contains triangle ACB, which is the same as triangle ABC.}\\
\cc{3. Question: Is point D the lies on line BC?}\\
\cc{Answer: \textbf{No}, point D is not on lines BC. The point D lies on the line segment AB.}\\
\cc{4. Question: Is line segment AC perpendicular to line segment CB?}\\
\cc{Answer: \textbf{Yes}, line segment AC is perpendicular to line segment CB.}\\
\end{tabular}
\end{sectionbox}
\vspace{-5mm}
\caption{Geometric image description and contrastive QA pairs for understanding basic elements. The generation process consists of two phases: 
1) Translating the human-labelled logic form into detailed information items and a summary of the diagram description.
2) Generating contrastive QA pairs based on the provided information and summary.
The sections \tcc{in blue} display the responses generated by ChatGPT. The detailed prompt will be included in the appendix.}
    \label{tab:qa_pairs_pretraining}
\end{minipage}
\end{table*}

\subsection{Geometric Cross-Modal Alignment Data}
\subsubsection{Geometric Image Caption Generation}
Image-caption datasets play a significant role in training MLLMs for understanding the context of images, which is essential for aligning image and text modalities. In the field of geometry, there is a lack of such datasets that offer detailed descriptions of geometric diagrams. To address this issue, we propose the generation of image descriptions from labeled question-answer (QA) pairs, as illustrated in Table~\ref{tab:generate_caption_pretraining}. In particular, we use text-only ChatGPT 3.5 to create image captions based on these human-labeled QA pairs, which can be considered as a type of inverse information recovery. This approach leverages the strong understanding ability of ChatGPT to produce descriptions for geometric diagrams. 

\subsubsection{Contrastive QA Pairs for Basic Elements}
Our approach also involves generating QA pairs to facilitate the comprehension of geometric diagrams, focusing primarily on their basic elements. The process begins with the interpretation of human-labeled logical forms on Geometry3k~\citep{lu2021inter}. We employ text-only ChatGPT to convert these logical forms into clear descriptions that cover various geometric elements such as shapes, lines, and points, and their relationships.

After creating these diagram descriptions, the model begins to produce contrastive QA pairs. These pairs are designed to examine different aspects of the diagrams. Questions may explore the presence of certain geometric elements (e.g., "Are there triangular shapes in the diagram?") or check the accuracy of the relationships described (e.g., "Is point D the lies on line BC?"). This method enables the model to comprehend geometric concepts and to analyze and interpret the details in geometric diagrams accurately.
The generation example is shown on Table~\ref{tab:qa_pairs_pretraining}.

\begin{table*}[t!]\centering
\begin{minipage}{1.0\textwidth}\vspace{0mm}    \centering
\begin{sectionbox}[]{Original Example}
    \centering
      \footnotesize
    \begin{tabular}{p{0.97\textwidth} c}
{ {\bf Question:} } & \\
As shown in the figure, the length of the short arm of the railing at the railway crossing is & \hspace{-3.2cm} \multirow{5}{*}{ \includegraphics[height=2.0cm]{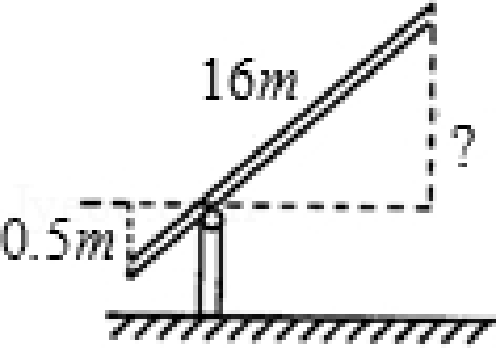} }\\
1.0 and the length of the long arm is 16.0. When the end of the short arm drops by 0.5, the \\
end of the long arm rises () \\
{\bf Answer: }  & \\
By the property of similar triangles, 
the lengths of corresponding sides are proportional.
Since \\the long arm is 16 times longer than the short arm, the height it rises is 16 times the height\\ the short arm drops, which is 0.5 multiplied by 16, resulting in a rise of 8 meters.
\\
\end{tabular}
\end{sectionbox}
\vspace{-5mm}
\caption{
The original example.}
    \label{tab:generated_examples}
\end{minipage}
\vspace{5mm}
\begin{minipage}{1.0\textwidth}\vspace{5mm}    \centering
\begin{sectionbox}[]{Value Scaling}
    \centering
      \footnotesize
    \begin{tabular}{p{0.97\textwidth} c}
{ {\bf Question:} } & \\
At a railroad crossing, the short arm of the barrier is \blue{2.0 unit} long and the long arm is \blue{32 units} & \hspace{-3.2cm} \multirow{5}{*}{ \includegraphics[height=2.0cm]{2350.png}  }\\
long. If the short arm drops \blue{1 units}, by how many units will the long arm rise?
\\
{\bf Answer: }  & \\
Denote the rise of the long arm as `h'. 
In similar triangles, the ratios of their corresponding \\
sides are in proportion.
h can be calculated using the equation 1/h = 2/32. Solving for `h' \\
gives h = 32 * 1 / 2, which equals 16 units. Therefore, the long arm rises by 16 units. \\
\end{tabular}
\end{sectionbox}
\vspace{-5mm}
\caption{
Synthetic example via value scaling.}
    \label{tab:value_scaling}
\end{minipage}
\end{table*}

\subsection{Geometric Instruction Data}
After performing alignment leveraging the constructed alignment data, the model is able to better interpret the geometric diagram (Figure~\ref{fig:comparison}). However, they are still limited at solving geometric problems. Therefore, we construct an instruction tuning dataset based on existing datasets with the help of powerful LLMs. Specifically, we design a series of strategies to expand the question-answer pairs in existing datasets. The resulting dataset contains more than 110k QA pairs, which is the largest public geometric QA dataset available. We will introduce the proposed strategies in detail below.

\subsubsection{Equation Solving (ES)} 
As shown in Table~\ref{tab:es}, we replace the specific values in the original QA pairs with unknown variables and prompt the LLM to construct the solution by solving equation. Such data is helpful for the MLLM to generalize its understanding of the problem, which enables it to apply the similar reasoning and solution steps to different scenarios. The abstraction of the problem by using variables and solving equation helps the LLM focus on the underlying mathematical concepts and relationships, rather than getting caught up in specific numerical values.

\subsubsection{Value Scaling (VS)} As shown in Table~\ref{tab:value_scaling}, we augment the data by scaling the length values in the QA pairs. Note that for the same diagram, the QA pair is still correct if all the lengths in a geometric problem are scaled simultaneously. However, note that it is not the case for quantities such as angles.  When different scaling of values are applied, the LLM becomes more flexible in handling different numerical inputs. Involving a range of values that extends beyond the initial training dataset aids in refining the model's computational and reasoning capabilities, thereby contributing to its generalizability.

\subsubsection{Re-Formulating Condition as Unknown (RCU)}
Motivated by\citep{weng2023large,yu2023metamath}, we design new multi-modal QA pairs that ask questions backwards, as shown in Table~\ref{tab:self_verification}. Specifically, we reformulate questions to ask for the values originally present in the condition, and retain the generated data with correct answer only. In this way, the LLM is repeatedly exposed to the relationships between variables, equations, and their solutions.
This reinforcement helps the model learn the dependencies and connections between different elements in a mathematical problem.

\subsubsection{Sentence Paraphrase (SP)}
We also conduct paraphrasing for both the question and answer pairs, as shown in Table~\ref{tab:paraphrase}. This exposes the LLM to a broader range of phrasing and language variations. This helps the model become more robust in understanding and generating diverse sentence structures. Consequently, it can handle similar questions with different phrasings and provide accurate responses.

\section{Model Architecture and Training}
\subsection{Model Architecture} We utilize the LLAVA~\citep{liu2023llava} architecture for our model.  The model mainly consists of a large language model (LLM) such as LLAMA-2~\cite{touvron2023llama}, a pretrained vision transformer~\cite{radford2021learning} (ViT) as image encoder. In addition, a projection layer is required to map the visual features from the image encoder to the same dimension as the LLM.

\begin{table*}[t!]\centering
\centering
\begin{sectionbox}[]{Variable Substitution and Equation Solving}
    \centering
      \footnotesize
    \begin{tabular}{p{0.97\textwidth} c}

{\bf Question: }  &  \hspace{-3.2cm} \multirow{5}{*}{ \includegraphics[height=2.0cm]{2350.png}  }\\
Given a railway crossing barrier with two arms, where the shorter arm has a length of \blue{`a'}\\ \blue{units (a = 1.0)} and the longer arm has a length of \blue{`b' units (b = 16.0)}, determine the elevation\\ \blue{`y' units} of the longer arm's end when the shorter arm descends by \blue{`d' units (d = 0.5).}\\ 
\blue{Express your answer as a function of `a', `b', and `d'.}
\\
\\
{\bf Answer: }  & \\
{To find the elevation `y' of the long arm's end, we apply the property of similar triangles and establish the equation \(\frac{d}{y} = \frac{a}{b}\). Substituting the given values, we get \(\frac{0.5}{y} = \frac{1.0}{16.0}\). Solving for 'y' leads to \(y = \frac{b \times d}{a}\), which simplifies to \(y = \frac{16.0 \times 0.5}{1.0}\). Therefore, 'y' equals 8.0 units, indicating that the long arm rises by 8.0 units when the short arm descends by 0.5 units.}

\rule{\linewidth}{1pt} & \\

{\bf Question: }  & \\
In a railway crossing lever system, the lengths of the shorter and longer arms are \blue{`s' units (s = 1.0)} and \blue{`l' units (l = 16.0)} respectively. If the shorter arm lowers by \blue{`deltaS' units (deltaS = 0.5),} \blue{what is the increase `deltaL' in the length of the longer arm?}
\\
\\
{\bf Answer: }  & \\
The increase `deltaL' in the longer arm can be found using the proportionality relationship of similar triangles. 
The equation can be expressed as: \(\frac{deltaS}{deltaL} = \frac{s}{l}\). Inserting the given values, the equation becomes \(\frac{0.5}{deltaL} = \frac{1.0}{16.0}\). Solving for `deltaL' yields \(deltaL = \frac{l \times deltaS}{s}\), which simplifies to \(deltaL = \frac{16.0 \times 0.5}{1.0}\). Thus, deltaL is 8.0 units.

\rule{\linewidth}{1pt} & \\

{ {\bf Question:} } & \\
At a railroad crossing, the short arm of the barrier is 1.0 unit long and the long arm is 16.0 units long. If the short arm drops 0.5 units, by how many units will the long arm rise? \\
\blue{Represent the rise of the long arm as `h'.}
\\
\\
{\bf Answer: }  & \\
The triangles formed by the shorter and longer arms as they descend are similar.
The rise of the long arm `h' can be calculated using 
the property of similar triangles, 
the equation is 0.5/h = 1/16. Solving for `h' gives h = 16 * 0.5 / 1, which equals 8.0 units. 
\end{tabular}
\end{sectionbox}
\caption{
Synthetic example via replacing values to unknown variables and solving the equation.}
    \label{tab:es}
\end{table*}

\begin{table*}[t!]\centering
\begin{minipage}{1.0\textwidth}\vspace{0mm}    \centering
\begin{sectionbox}[]{Re-Formulating Condition as Unknown~\citep{weng2023large,yu2023metamath}}
    \centering
      \footnotesize
    \begin{tabular}{p{0.97\textwidth} c}
{ {\bf Question:} } & \\
At a railroad crossing, the short arm of the barrier is \blue{u} unit long and the long arm is 16.0 & \hspace{-3.2cm} \multirow{5}{*}{ \includegraphics[height=2.0cm]{2350.png}  }\\
units long. When the end of the short arm drops by 0.5, the end of the long arm rises 8 units. 
\\ \blue{What is the value of unknown variable u?}
\\
\\
{\bf Answer: }  & \\
Denote the short arm of the barrier as variable u. By the property of similar triangles,\\ we can determine u by solving the equation 0.5/8 = u/16. Therefore, u equals 1. \\
\end{tabular}
\end{sectionbox}
\vspace{-2mm}
\caption{
Synthetic example via re-formulating condition as unknown.}
    \label{tab:self_verification}
\end{minipage}
\end{table*}

\begin{table*}[t!]\centering
\begin{minipage}{1.0\textwidth}\vspace{0mm}    \centering
\begin{sectionbox}[]{Sentence Paraphrase}
    \centering
      \footnotesize
    \begin{tabular}{p{0.97\textwidth} c}
{ {\bf Question:} } & \\
In the illustration, the railing at the railway crossing has a short arm measuring 1.0 unit & \hspace{-3.2cm} \multirow{5}{*}{ \includegraphics[height=2.0cm]{2350.png}  }\\
in length and a long arm measuring 16.0 units. When the short arm drops by 0.5 units,\\ what is the corresponding rise in the long arm?
\\
\\
{\bf Answer: }  & \\
The triangles are similar, and their corresponding sides are proportional. The long arm \\ is 16 times longer than the short arm, resulting in an 8-meter rise when the short arm \\drops by 0.5 meters. \\
\end{tabular}
\end{sectionbox}
\vspace{-2mm}
\caption{
Synthetic example via sentence paraphrase.}
    \label{tab:paraphrase}
\end{minipage}
\end{table*}

During inference, given an image and a textual instruction, the image encoder first extracts the visual tokens from the image, which are then mapped to the dimension of LLM's embedding space via the projection layer. Then, the mapped image features are concatenated with text embeddings to serve as the input to the LLM. Subsequently, the LLM begins to perform next-token-generation.

\subsection{Model Training} We train our \ourmodel in two phases, namely 1) geometric visual-language alignment, and 2) geometric instruction tuning. In both phases, we leverage the conventional language modeling loss, which can be formulated as follows:

\begin{small}
\begin{align}
    \mathcal{L}(S_{tar}, S_{in}, I)=-\sum^L_{t=1}\log p\left[S^t_{tar} | \mathcal{F} (s^{(<t)}_{tar}, S_{in}, I)\right]
\end{align}
\end{small}

\hspace{-4mm}where $\mathcal{F}$ represents the model. $I$ represents the geometric figure; $S_{tar}$ and $S_{in}$ represent the target and input sentences, respectively; $S^t_{tar}$ denotes the $t^{th}$ token of target output, and $L$ stands for length. 
\section{Experiments}
\subsection{Setup}
\begin{table}[t!]
\centering
\resizebox{\linewidth}{!}{
\begin{tabular}{l|c|c}
\toprule
\textbf{Model} &\textbf{Input}& \textbf{Accuracy (\%)} \\ \hline
\multicolumn{3}{c}{\textit{Heuristics Baseline}} \\
Random Chance & -  & 21.6 \\ 
Frequent Guess &- & 34.1 \\ 
\bb{{Human}} &\bb{{$Q,I$}} & \bb{{48.4}} \\
\midrule
\multicolumn{3}{c}{{{\textit{Close Source Model}}}} \\
\multicolumn{3}{l}{\small{\textbf{\textit{Text-Only LLMs}}}} \\
2-shot CoT Claude-2  & $Q$ & 29.8 \\ 
2-shot CoT ChatGPT  & $Q$  &  36.5\\ 
2-shot CoT GPT-4  &$Q$ & 44.7 \\ 
2-shot PoT ChatGPT  & $Q$ &30.8  \\ 
2-shot PoT GPT-4  &$Q$ & 33.2
 \\
\multicolumn{3}{l}{\textbf{\small{\textit{Visual-Augmented LLMs}}}} \\
2-shot CoT Claude-2  & $Q,I_{c}, I_{t}$ & 31.7 \\ 
2-shot CoT ChatGPT  & $Q,I_{c}, I_{t}$  & 29.3 \\ 
2-shot CoT GPT-4  &$Q,I_{c}, I_{t}$ & 31.7 \\ 
2-shot PoT ChatGPT  & $Q,I_{c}, I_{t}$ & 26.4 \\ 
2-shot PoT GPT-4  &$Q,I_{c}, I_{t}$ & 39.4 \\
\multicolumn{3}{l}{\textbf{\small{\textit{Multimodal LLMs}}}} \\
Multimodal Bard &$Q,I$ & 47.1 \\ 
Gemini Nano 1 & $Q,I$ &21.6\\
Gemini Nano 2 & $Q,I$ &23.6\\
Gemini Pro&$Q,I$ & 40.4\\
\bb{Gemini Ultra}&\bb{$Q,I$} & \bb{56.3}\\
\bb{{{GPT4-V}}} &\bb{{{$Q,I$}}}& \bb{{{50.5}}} \\ 
\midrule
\multicolumn{3}{c}{{{\textit{Open Source Model}}}} \\
IDEFICS (9B-Instruct) & $Q,I$ & 21.1 \\
mPLUG-Owl (LLaMA-7B)& $Q,I$ & 23.6 \\ 
miniGPT4 (LLaMA-2-7B)  & $Q,I$ & 26.0 \\
LLaMA-Adapter-V2 (7B)  & $Q,I$ & 25.5 \\ 
LLaVAR & $Q,I$ & 25.0 \\ 
InstructBLIP (Vicuna-7B)  & $Q,I$ & 20.7 \\ 
LLaVA (LLaMA-2-13B) & $Q,I$ & 29.3 \\
\cc{{\ourmodel-7B} }&\cc{$Q,I$ }& \cc{{53.4}} \\ 
\cc{\textbf{\ourmodel-13B }}& \cc{$Q,I$} & \cc{\textbf{56.7}} \\ 
\bottomrule
\end{tabular}
}
\caption{Comparison of model performance on the testmini set of MathVista benchmarks~\citep{lu2023mathvista} on geometry problem solving (GPS) . For input, $Q$ represents for question, $I$ represents for image, $I_c$ represents for image caption generated by Bard, and $I_t$ represents fo OCR text detected in the image. Baseline results are obtained from \citet{lu2023mathvista}. 
Human performance and the results surpassing human performance are highlighted in \hl{grey}. Our results are highlighted in \colorbox{lightblue}{blue}.
} 
\label{tab:model_performance_mathvista}
\end{table}

\paragraph{Dataset.}
We generate the alignment data and instruction data utilizing training set of GeoQA+~\citep{cao2022augmented} and Geometry3K~\citep{lu2021inter}. More specifically, the contrastive question-answer (QA) pairs in the alignment data are generated using Geometry3K, which features human-labeled logical forms.
Note that GeoQA+ covers the training set of GeoQA~\citep{chen-etal-2021-geoqa}, and share the same val/test set as GeoQA~\citep{chen-etal-2021-geoqa}.
More details of data split on GeoQA and GeoQA+ is listed in Table~\ref{tab:dataset_split}. 
Our approach results in 60K alignment data samples, and more than 110K instruction data samples.

We compare our model with other MLLMs on the geometry problems on the minitest split MathVista \citep{lu2023mathvista}, and compare our model with traditional in-domain model on the test split of GeoQA following~\citep{chen2022unigeo,liang_unimath}. The geometry problems in MathVista minitest set is collected from four source datasets Geometry3K~\citep{lu2021inter}, GeoQA+~\citep{cao2022augmented}, GEOS~\citep{seo2015solving} and UniGeo~\citep{chen2022unigeo}. 
\begin{table}[h]
\centering
\resizebox{\linewidth}{!}{
\begin{tabular}{l|c|c|c}
\toprule
Dataset & Train & Validation & Test \\ \hline
GeoQA+~\citep{cao2022augmented} & 6027 & 745 & 754 \\
GeoQA~\citep{chen-etal-2021-geoqa} & 3499 & 745 & 754 \\ 
\bottomrule
\end{tabular}
}
\vspace{-2mm}
\caption{Data Split of GeoQA and GeoQA+.}
\vspace{-3mm}
\label{tab:dataset_split}
\end{table}

\begin{table}[h]
\centering
\resizebox{\linewidth}{!}{
\begin{tabular}{l|c|c}
\toprule
\textbf{Model} &\textbf{Input} & \textbf{Accuracy (\%)} \\ \hline
Random Chance & - &25.0  \\ 
Frequent Guess &- & 32.1  \\ 
\hline
\multicolumn{3}{c}{\textit{Top-10 Accuracy}} \\
NGS~\citep{chen-etal-2021-geoqa} & $Q,I$ & 56.9\\
DPE-GPS~\citep{cao2022augmented} & $Q,I$& 62.7\\
SCA-GPS~\citep{sca_gps} & $Q,I$& 64.1\\
\hline
\multicolumn{3}{c}{\textit{Top-1 Accuracy}} \\
Geoformer~\citep{chen2022unigeo} & $Q,I$ & 46.8\\
UniMath~\cite{liang_unimath}& $Q,I$ & 50.0\\
\cc{\textbf{\ourmodel-7B}} &\cc{$Q,I$} &  \cc{\textbf{64.2 }}\\ 
\cc{\textbf{\ourmodel-13B}}& \cc{$Q,I$}&   \cc{\textbf{{67.0}}}\\ 
\bottomrule
\end{tabular}
}
\vspace{-2mm}
\caption{Comparison of model performance with traditional methods on GeoQA.}
\label{tab:model_performance_geoqa}
\end{table}

\paragraph{Implementation Details.} 
We employ ChatGPT (gpt-3.5-turbo-0613) for data generation. A detailed description of our prompts will be provided in the appendix. 
We use LLaVA~\citep{liu2023llava} as our backbone.
More specifically, we utilize LLAMA-2~\citep{touvron2023llama} as the language model and employ the visual encoder of a pretrained vision transformer~\cite{radford2021learning} (ViT). The resolution of the input image is 336 by 336. We conduct experiments with both 7B and 13B LLMs. 
In the cross-modal alignment process, only the projection linear layer is trainable.  During the instruction tuning phase, both the projection linear layer and the language model are trainable.

For training data, as we found the minitest split of MathVista contains some examples of Mix-train.pk of GeoQA+, we remove those samples that also appears in minitest split of MathVista. 
The learning rate is set to $3e^{-5}$. We expand the images into squares during training, where the extended background color is set to white. For image augmentation, we set the maximum translation distance to 0.25 of the length of longer side. 
If not otherwise specified, the models are trained for 1 epoch for cross-modal alignment and 2 epochs for instruction tuning, respectively. And the batch sizes are set to 6 per GPUs and 32 per GPUs, respectively.

\paragraph{Evaluation Metric.}
We use accuracy as the metric for evaluation. 
Note that several prior studies \citep{chen-etal-2021-geoqa,chen2022unigeo,cao2022augmented} report results using Top-10 accuracy (generating 10 sequences and selecting the first sequence that successfully addresses the problem as the prediction).  
Our experimental results directly report Top-1 accuracy.
During instruction tuning, we enable the model to output the choice in a fixed format. For evaluation, we directly use regular expression to extract the predicted choices from the generated answers. The answer is considered false if the regular expression fails to extract a valid answer.

\subsection{Main Experiment}

We compared \ourmodel with other MLLMs on minitest split of MathVista~\citep{lu2023mathvista} benchmark on Table~\ref{tab:model_performance_mathvista}. The results shows that, geometric cross-modal alignment and instructing tuning on our dataset is effective in improve MLLMs' geometric problem solving ability. Our specific in-domain model \ourmodel-7B can even surpass the strong GPT4-V on geometric problems.

\subsection{Comparison with Conventional Methods}
We additionally compare our method with conventional SOTA methods in geometry problem solving domain. As illustrated in Table~\ref{tab:model_performance_geoqa}, our method demonstrates a notable improvement in Top-1 accuracy over the existing SOTA techniques. Moreover, our model's top-1 accuracy outperforms the baselines' top-10 accuracy, demonstrating a significant improvement in predictive precision.

\subsection{Performance Across Problem Difficulties}
We compare \ourmodel with the baselines models on  problems with different difficulty levels, as shown in Table~\ref{tab:difficulty}. Specifically, OP represents the number of ``operations", or reasoning steps that needs to be taken for solving the problem. The results verify that our \ourmodel consistently outperforms baseline models by a large margin across various difficulty levels.
\begin{table}[h]
    \centering
    \resizebox{\linewidth}{!}{
    \scalebox{0.70}{
    \begin{tabular}{l|c|c|c|c|c}
    \toprule
    \textbf{Model} & \textbf{OP=1(\%)} & \textbf{OP=2(\%)} & \textbf{OP=3(\%)} & \textbf{OP\textgreater=4(\%)} & \textbf{Total(\%)} \\
    \hline
    LLaVA-7B &16.8 &20.9 & 15.5&22.9&18.7\\
    LLaVA-13B &19.1 &21.3 & 18.5&24.6 & 20.3 \\
    \textbf{\cc{\ourmodel-7B}} & \textbf{\cc{77.5}} & \textbf{\cc{60.8}} & \textbf{\cc{54.8}} & \textbf{\cc{40.9}} & \textbf{\cc{64.2}} \\
    \textbf{\cc{\ourmodel-13B}} & \textbf{\cc{79.0} }& \textbf{\cc{64.9}} & \textbf{\cc{55.5}} & \textbf{\cc{49.1} }& \textbf{\cc{67.0}} \\
    \bottomrule
    \end{tabular}
    }
    }
    \vspace{-2mm}
    \caption{Different difficulty problems on GeoQA.}
    \vspace{-2mm}
    \label{tab:difficulty}
\end{table}

\vspace{-1em}
\subsection{Performance Across Different Types of Questions}
We compare \ourmodel with the baselines models on  problems with different type of questions, as shown in Table~\ref{tab:diff_types_problems}. The results suggest that \ourmodel performs better than the baseline models in various geometric problems such as angle, length, and area problems.
\begin{table}[h]
\centering
\resizebox{\linewidth}{!}{
\begin{tabular}{l|c|c|c|c|c}
\toprule
\textbf{Model} & \textbf{Angel} & \textbf{Length} & \textbf{Area} & \textbf{Other} & \textbf{Total} \\ \hline
LLaVA-7B &16.1&22.2&17.0&14.3&18.7\\
LLaVA-13B &17.5&23.0&25.5&28.6&20.3\\
\textbf{\cc{\ourmodel-7B}}& \textbf{\cc{70.7}}& \textbf{\cc{56.5}}& \textbf{\cc{55.3}}& \textbf{\cc{42.9}}& \textbf{\cc{64.2}}\\ 
\textbf{\cc{\ourmodel-13B}} & \textbf{\cc{71.5}}& \textbf{\cc{61.1}}& \textbf{\cc{63.8}}& \textbf{\cc{57.1}}& \textbf{\cc{67.0}}\\ 
\bottomrule
\end{tabular}
}
\vspace{-2mm}
\caption{Performance of different types of questions on GeoQA.}
\vspace{-2mm}
\label{tab:diff_types_problems}
\end{table}
\begin{table}[b]
\centering
\resizebox{\linewidth}{!}{
\begin{tabular}{l|c|c}
\toprule
\textbf{Model} &\textbf{Input} & \textbf{Accuracy (\%)} \\ \hline
Random Chance & - &25.0  \\ 
Frequent Guess &- & 32.1  \\ 
\hline
Geoformer~\citep{chen2022unigeo} & $Q,I$ & 46.8\\
UniMath~\cite{liang_unimath}& $Q,I$ & 50.0\\
\cc{\textbf{\ourmodel-7B}} &\cc{$Q,I$} &  \cc{\textbf{64.2 }}\\ 
\cc{\quad w/o alignment phase}& \cc{$Q,I$}& \cc{{62.8}}\\ 
\bottomrule
\end{tabular}
}
\caption{Effectiveness of alignment in the pre-training phase. Top-1 accuracy is reported.}
\label{tab:geoqa_pretraining}
\end{table}

\begin{figure*}[t!]
\centering
\includegraphics[width=1\textwidth]{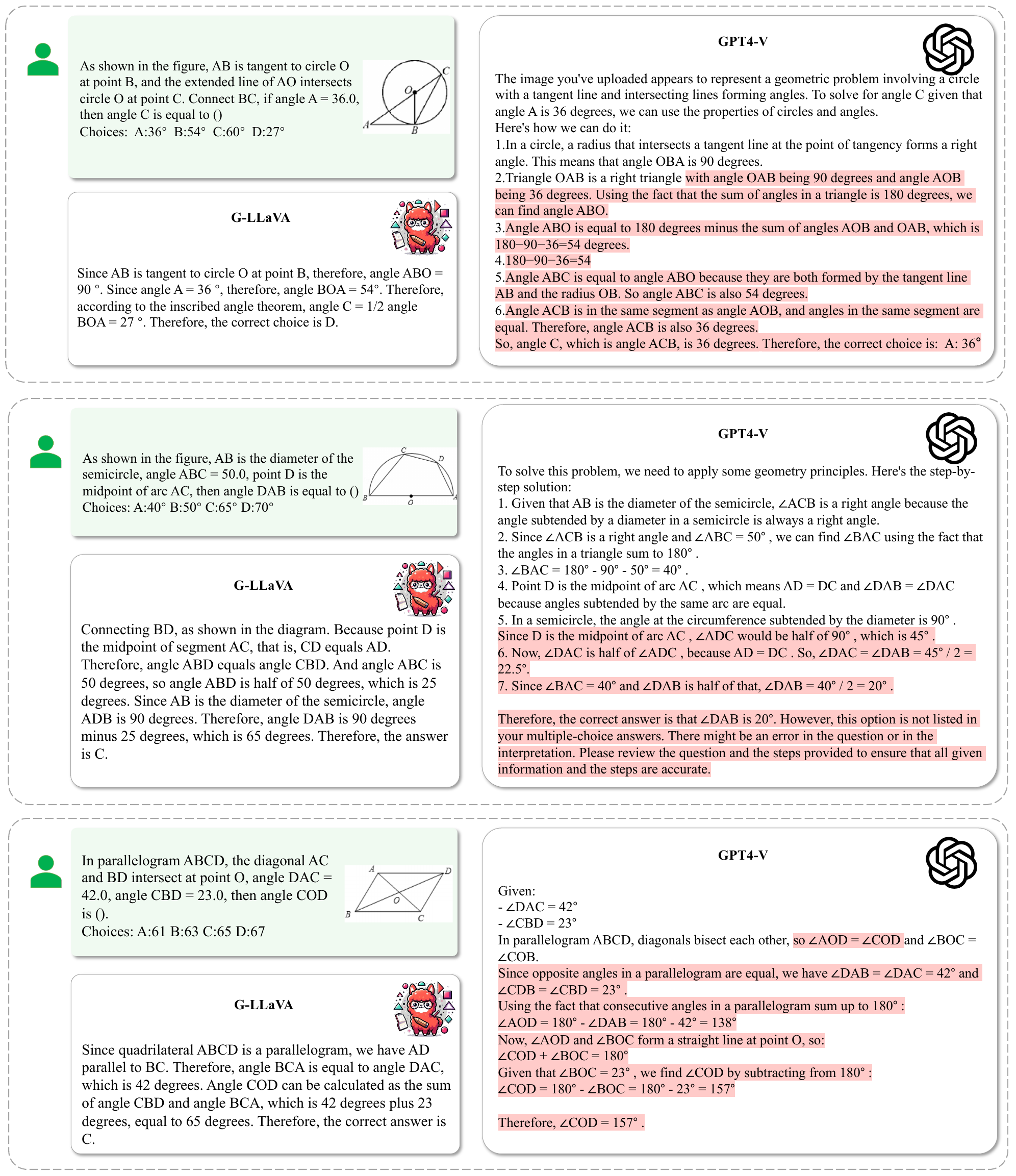} 
\caption{Demonstration of geometric problem solving by GPT-4-V and \ourmodel.
}\label{fig:demo}
\end{figure*}

\vspace{-1em}
\subsection{Effectiveness of Cross-Modal Geometric Alignment}
To evaluate the alignment phase's effectiveness, we conducted the analysis of the model's performance with and without alignment  phase in Table~\ref{tab:geoqa_pretraining}. The results suggest that the alignment phase enhances the model's ability to interpret images, which is also illustrated
by the qualitative result 
in Figure~\ref{fig:comparison}.

\section{Conclusion}
In this paper, we make the attempt to address the limitations of current MLLMs in solving geometric problems. We propose strategies to enrich the data by leveraging LLMs, resulting in our augmented dataset, \ourdata. With this dataset, our \ourmodel outperforms GPT-4-V on the geometric split of MathVista benchmark, with as few as 7B parameters. We hope our work provides new insights on improving multimodal LLMs' ability of solving geometric problems.

\section{Acknowledgement}

We would like to thank Zhenwen Liang for his valuable discussions and insightful feedback.

\bibliography{anthology,custom}
\bibliographystyle{acl_natbib}
\appendix

\end{document}